\pdfoutput=1

\documentclass[11pt]{article}

\usepackage{ACL2023}

\usepackage{times}
\usepackage{latexsym}

\usepackage[T1]{fontenc}

\usepackage[utf8]{inputenc}

\usepackage{microtype}

\usepackage{inconsolata}

\usepackage{arydshln}
\usepackage{multirow}
\usepackage{amsmath}
\usepackage{CJKutf8}
\usepackage{graphicx}
\usepackage{amssymb}

\newcommand{\bhline}{\noalign{\hrule height 1.0pt}}

\title{Acquiring Frame Element Knowledge with Deep Metric Learning \\ 
for Semantic Frame Induction}

\author{
  Kosuke Yamada$^{1}$ \ \ \ \ \ \ \ \ \ 
  Ryohei Sasano$^{1,2}$ \ \ \ \ \ \ \ \ \ 
  Koichi Takeda$^{1}$  \\
  $^{1}$Graduate School of Informatics, Nagoya University, Japan \\
  $^{2}$RIKEN Center for Advanced Intelligence Project, Japan \\
  {\tt yamada.kosuke.v1@s.mail.nagoya-u.ac.jp}, \\
  {\tt \{sasano,takedasu\}@i.nagoya-u.ac.jp}
}

\begin{document}
\maketitle
\begin{abstract}
The semantic frame induction tasks are defined as a clustering of words into the frames that they evoke, and a clustering of their arguments according to the frame element roles that they should fill.
In this paper, we address the latter task of argument clustering, which aims to acquire frame element knowledge, and propose a method that applies deep metric learning.
In this method, a pre-trained language model is fine-tuned to be suitable for distinguishing frame element roles through the use of frame-annotated data, and argument clustering is performed with embeddings obtained from the fine-tuned model.
Experimental results on FrameNet demonstrate that our method achieves substantially better performance than existing methods.
\end{abstract}

\section{Introduction}
A semantic frame is a coherent conceptual structure that describes a particular type of situation or event along with its participants and props.
FrameNet \cite{ruppenhofer-2016-framenet} is a representative resource, in which semantic frames define a set of frame-specific roles called frame elements (FEs).
FrameNet comprises a list of semantic frames, sets of frame-evoking words, and collections of frame-annotated examples.
Table \ref{tab:examples} lists examples of frame-annotated sentences for the \textsc{Giving} frame in FrameNet.
For each sentence, a frame-evoking word is annotated with the \textsc{Giving} frame, and its arguments are annotated with FEs such as \textsf{\small Donor}, \textsf{\small Theme}, and \textsf{\small Recipient}.

Because manually arranging such frame resources on a large scale is labor intensive, there have been many studies on automatic induction of frame resources.
Most of these studies have assumed only verbs as frame-evoking words and divided the frame induction task into two sub-tasks: verb clustering, which groups verbs according to the frames that they evoke, and argument clustering, which groups arguments of verbs according to their FE roles \cite{anwar-2019-hhmm,ribeiro-2019-l2f}.
This study addresses the argument clustering task and acquires frame element knowledge for semantic frame induction.

\begin{table}[!t]
\centering
\begin{tabular}{@{\ }l@{\ }} \bhline
\rule{0ex}{2.5ex}
1. $[_{(1){\color[HTML]{ff7f0e} \rm \scriptsize \sf Theme}}$ It$]$ was \underline{handed} in $[_{(2){\color[HTML]{1f77b4} \rm \scriptsize \sf Donor}}$ by a \\
\ \ \ \ \ couple of children$]$  this morning. \\
\rule{0ex}{2.0ex}
2. $[_{(3){\color[HTML]{1f77b4} \rm \scriptsize \sf Donor}}$ I$]$ will now \underline{donate} $[_{(4){\color[HTML]{ff7f0e} \rm \scriptsize \sf Theme}}$ the \\
\ \ \ \ \ money$]$ $[_{(5)}$$_{\color[HTML]{2ca02c} \rm \scriptsize \sf Recipient}$ to charity$]$. \\
\rule{0ex}{2.0ex}
3. $[_{(6){\color[HTML]{1f77b4} \rm \scriptsize \sf Donor}}$ Your gift$]$ \underline{gives} $[_{(7){\color[HTML]{2ca02c} \rm \scriptsize \sf Recipient}}$ children \\
\ \ \ \ \ and families$]$ $[_{(8){\color[HTML]{ff7f0e} \rm \scriptsize \sf Theme}}$ hope for tomorrows$]$. \\ \bhline
\end{tabular}
\caption{Examples of verbs that evoke the \textsc{Giving} frame in FrameNet}
\label{tab:examples}
\end{table}

\begin{figure}[!t]
\centering
\includegraphics[width=\linewidth]{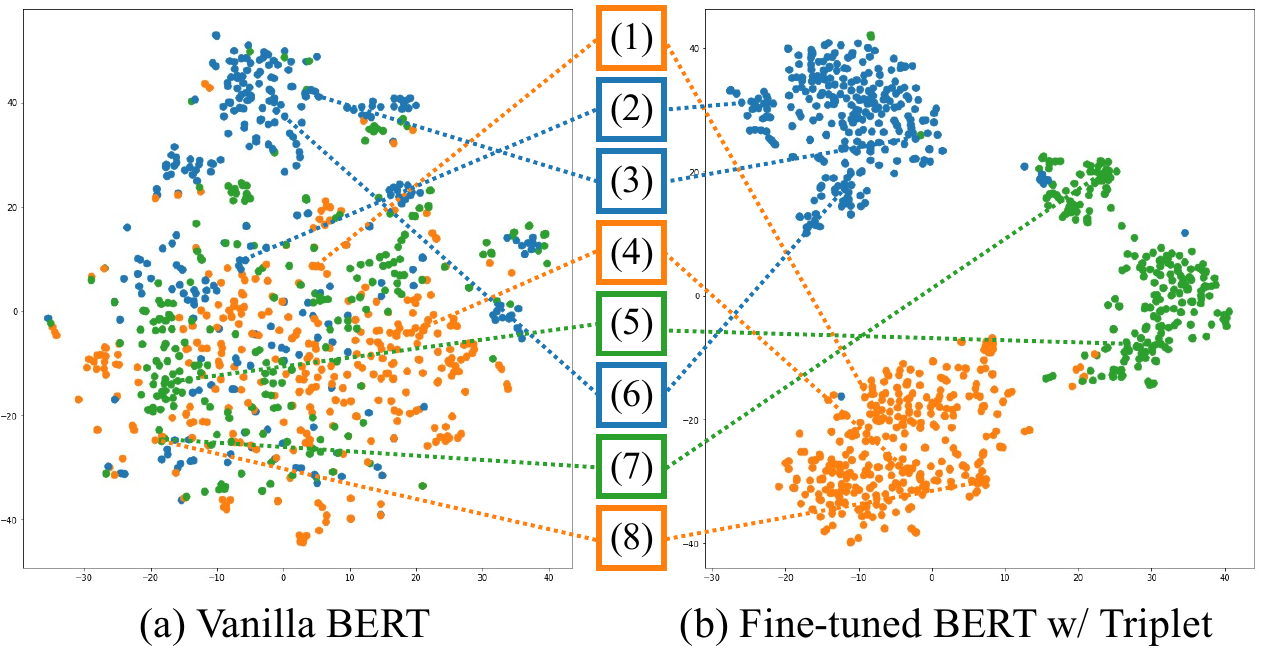}
\caption{2D t-SNE mappings of average BERT embeddings of argument tokens, which are labeled with {\color[HTML]{1f77b4} \textsf{\small Donor}}, {\color[HTML]{ff7f0e} \textsf{\small Theme}}, or {\color[HTML]{2ca02c} \textsf{\small Recipient}}, in examples of verbs that evoke the \textsc{Giving} frames in FrameNet. 
The numbers in parentheses correspond to the examples in Table \ref{tab:examples}.} 
\label{fig:examples}
\end{figure}

As with many natural language processing tasks, methods using contextualized embeddings such as ELMo \cite{peters-2018-deep} and BERT \cite{devlin-2019-bert} have been proposed for argument clustering tasks.
However, these methods have been reported to perform worse than methods based on syntactic relations \cite{anwar-2019-hhmm,ribeiro-2019-l2f}.
We assume that this is because vanilla BERT, i.e., BERT without fine-tuning, is more influenced by factors such as a whole sentence's meaning and does not emphasize information that captures differences in semantic roles.
Figure \ref{fig:examples}(a) shows a 2D t-SNE \cite{maaten-2008-visualizing} projection of the average BERT embeddings of argument tokens in examples of the \textsc{Giving} frame in FrameNet.
We can see that these embeddings are not adequately clustered according to their semantic roles.

Hence, in this study, we propose the use of deep metric learning to fine-tune a contextual word embedding model so that instances of the same FEs are placed close together while other instances are placed farther apart in the embedding space.
Figure \ref{fig:examples}(b) shows a 2D projection of the average BERT embeddings of argument tokens after fine-tuning with our proposed method based on the triplet loss.
We can confirm that instances of the same FEs are located close to each other.
This suggests that deep metric learning enables fine-tuning of BERT to obtain embedding spaces that better reflect human intuition about FEs.

\section{Acquiring Frame Element Knowledge with Deep Metric Learning}
To acquire frame element knowledge for semantic frame induction, we work on argument clustering, which is the task of grouping arguments of frame-evoking words according to their roles in the evoked frame.
We introduce two argument clustering methods that cluster argument instances using their contextualized word embeddings.
To achieve higher performance methods, we assume the existence of frame-annotated data and propose to fine-tune a contextualized word embedding model using deep metric learning.

\subsection{Deep Metric Learning}
Deep metric learning is a method of learning deep learning models on the embedding space in such a way that instances with the same label are placed closer together and instances with different labels are placed farther apart \cite{kaya-2019-deep,musgrave-2020-metric}.
By applying this to the contextualized word embedding model, it is expected that argument instances with similar roles learn to be closer together, and argument instances with different roles learn to be farther apart.
We use the representative triplet \cite{weinberger-2009-distance} and ArcFace losses \cite{deng-2019-arcface} from two major approaches: the distance-based and classification-based approaches, respectively.

\paragraph{Triplet loss}
This loss function is commonly used in deep metric learning, in which the distance to a triplet of instances can be learned directly using three encoders.
Specifically, it performs learning such that the distance between an anchor instance $\boldsymbol{x}_a$ and a negative instance $\boldsymbol{x}_n$, which are taken from different classes, is to be larger than a certain margin $m$ plus the distance between the anchor instance $\boldsymbol{x}_a$ and a positive instance $\boldsymbol{x}_p$.
The squared Euclidean distance is typically used as the distance function $D$.
The triplet loss is defined as follows:
\begin{equation}
\!\!\! L_{\mathrm{tri}} \! = \!
\max \left(D\left(\boldsymbol{x}_a, \boldsymbol{x}_p \right)
\! - \! D\left(\boldsymbol{x}_a, \boldsymbol{x}_n \right)
\! + \! m,
0 \right).
\label{eq:triplet}
\end{equation}

\paragraph{ArcFace loss}
This loss has been used as a de facto standard in face recognition.
It modifies the softmax-based cross-entropy loss for typical $n$-class classifiers.
Specifically, it applies $l_2$ regularization to the $i$-th class weight $\boldsymbol{w}_i$ and the embedding of the $i$-th class instance $\boldsymbol{x}_i$.
The angle between $\boldsymbol{w}_i$ and $\boldsymbol{x}_i$ is denoted as $\theta_i$.
An angular margin $m$ and a feature scale $s$ are introduced as hyperparameters to simultaneously enhance the intra-class compactness and inter-class discrepancy.
The ArcFace loss is defined as follows:
\begin{equation}
\!\!\! L_{\mathrm{arc}} \! = \!
- \log \frac{e^{s \cdot \cos \left(\theta_i+m \right)}}
{e^{s \cdot \cos \left(\theta_i+m \right)}
\! + \! \sum_{j=1, j \neq i}^{n} e^{s \cdot \cos \theta_j}}.
\label{eq:arcface}
\end{equation}

\subsection{Argument Clustering Methods}
We introduce two argument clustering methods: a cross-frame clustering of argument instances across frames and an intra-frame clustering of frame-wise argument instances.

\subsubsection{Cross-Frame Method}
The cross-frame method is a method used by \newcite{anwar-2019-hhmm} and \newcite{ribeiro-2019-l2f}, in which FEs are regarded as general semantic roles independent of frames, and the argument instances are grouped by roles across frames.
For example, both \textsf{\small Donor} in the \textsc{Giving} frame and \textsf{\small Agent} in the \textsc{Placing} frame are similar roles in the meaning of ``a person who acts on an object.''
Taking advantage of this property, the cross-frame method clusters the argument instances to form role clusters without considering the frame that each word evokes and then combines the frame and the role clusters into the FE clusters.
In this method, we apply group-average clustering based on the Euclidean distance, which is a hierarchical clustering algorithm.\footnote{See Appendix \ref{sec:number_of_clusters} for the number of clusters.}

The cross-frame method performs fine-tuning of contextualized word embedding models across frames by using the triplet and ArcFace losses.
For the triplet loss, a positive instance is one with the same FE as the anchor instance, while a negative instance is one with FEs of different frames or different FEs of the same frame as the anchor instance. 
The ArcFace loss is used to classify instances on an FE basis so that the model trains the metric across frames rather than within a particular frame.

\subsubsection{Intra-Frame Method}
Since the cross-frame method treats FEs as roles independent of frames even though FEs are frame-specific roles, there are two possible drawbacks as described below.
We thus propose the intra-frame method that treats FEs as frame-specific roles.

As the first drawback, the cross-frame method causes the division of argument instances of the same FE into too many clusters.
For example, the \textsc{Giving} frame has only three FEs, but the cross-frame method is likely to split instances into more clusters due to the nature of clustering across frames.
To overcome this drawback, the intra-frame method focuses on clustering the argument instances for each frame.
The method also uses group-average clustering.

As the second drawback, the fine-tuning of the cross-frame method may not provide the optimal embedding space for argument roles, because it learns to keep instances with similar roles in different frames away from each other.
For example, \textsf{\small Donor} in the \textsc{Giving} frame and \textsf{\small Agent} in the \textsc{Placing} frame are similar, but the cross-frame method keeps these instances away because they are regarded as different roles.
Hence, the intra-frame method learns to keep away only between instances of different FEs of the same frame.
For the triplet loss, this is achieved by limiting negative instances to be different FEs in the same frame.
For the ArcFace loss, this is achieved by training classification for the number of FE types in a frame.

\section{Experiment}
To confirm the usefulness of fine-tuning with deep metric learning, we experimented with an argument clustering task.
This study focuses on argument clustering to induce FEs for frame-evoking verbs.
Given the true frame that a verb evokes and the true positions of its argument tokens in the example sentences, we cluster only its arguments to generate role clusters. 
Then, we merge the true frame and the role clusters to obtain the final FE clusters.

\subsection{Settings}
\paragraph{Dataset}
The dataset in our experiment was created by extracting example sentences, in which the frame-evoking word was a verb, from FrameNet 1.7.\footnote{\url{https://framenet.icsi.berkeley.edu/}}
The FEs in FrameNet are divided into two types: core FEs, which are essential for frames, and non-core FEs.
Our experiment targeted only the core FEs, as in \newcite{qasemizadeh-2019-semeval}.
The examples were divided into three sets so that those of the verbs that evoke the same frames were in the same set.
Table \ref{tab:dataset} lists the dataset statistics.
We performed three-fold cross-validation with the three sets as the training, development, and test sets.
Note that the frames to be trained and those to be clustered do not overlap because the sets are divided on the basis of frames.

\begin{table}[!t]
\centering
\begin{tabular}{@{\ }l@{\ \ }rr@{\ \ \ }r@{\ \ \ }r@{\ }} \bhline
& \#Frames  & \#FEs & \#Examples & \#Instances \\ \bhline
Set 1   & 212  &   641 & 21,433 &  42,544 \\ 
Set 2   & 212  &   623 & 24,582 &  47,629 \\ 
Set 3   & 213  &   637 & 35,468 &  71,617 \\ \hline
All     & 637  & 1,901 & 81,493 & 161,790 \\ \bhline
\end{tabular}
\caption{Statistics of the FrameNet-based dataset used in three-fold cross-validation.}
\label{tab:dataset}
\end{table}

\paragraph{Comparison Methods}
We used BERT\footnote{\url{https://huggingface.co/bert-base-uncased}} from Hugging Face \cite{wolf-2020-transformers} to obtain contextualized word embeddings.
We compared a total of six different methods, which use the cross-frame method or the intra-frame method for each of the three models, the vanilla model (\textbf{Vanilla}) and two fine-tuned models (\textbf{Triplet}, \textbf{ArcFace}).\footnote{See Appendix \ref{sec:detail_settings} for the detailed settings of these methods.}

\begin{table*}[t!]
\centering
\begin{tabular}{ll ccc} \bhline
\multicolumn{2}{l}{Method} & \#C & \textsc{Pu} / \textsc{iPu} / \textsc{PiF}  & \textsc{BcP} / \textsc{BcR} / \textsc{BcF} \\ \bhline
\rule{-0.5ex}{2.0ex}
Boolean && 411 & 70.7 / 85.9 / 77.6 & 61.4 / 79.6 / 69.4 \\
Dependency-relationship && 2,032 & 84.6 / 70.6 / 77.0 & 78.2 / 56.9 / 65.9 \\ \hline
\rule{-0.7ex}{2.0ex}
\newcite{anwar-2019-hhmm} && 415 & 59.2 / 75.8 / 66.5 & 49.0 / 67.0 / 56.6 \\ 
\newcite{ribeiro-2019-l2f} && 628 & 65.3 / 74.6 / 69.6 & 55.0 / 64.4 / 59.3 \\ \bhline 
{\small \textbf{\underline{Clustering}}} & {\small \textbf{\underline{Model}}} & & & \\

\multirow{2.2}{*}{Cross-frame method}
& Vanilla  & 628 & 55.2 / 87.5 / 67.6 & 46.5 / 81.1 / 59.0 \\ \cdashline{2-5}
\rule{0ex}{2.2ex}
\multirow{1.8}{*}{(group-average clustering)}
& Triplet  & 543 & 80.0 / 92.9 / 86.0 & 73.0 / 88.8 / 80.1 \\
& ArcFace  & 594 & 81.7 / 91.5 / 86.2 & 74.9 / 86.8 / 80.3 \\ \hline 

\rule{0ex}{2ex}
\multirow{2.2}{*}{Intra-frame method}
& Vanilla  & 636 & 54.9 / 88.9 / 67.9 & 46.2 / 83.1 / 59.4 \\ \cdashline{2-5}
\rule{0ex}{2.2ex}
\multirow{1.8}{*}{(group-average clustering)}
& Triplet  & 646 & \textbf{90.1} / \textbf{95.0} / \textbf{92.5} & \textbf{85.5} / \textbf{91.9} / \textbf{88.6} \\
& ArcFace  & 631 & 90.0 / 94.3 / 92.1 & 85.4 / 90.9 / 88.1 \\ \bhline
\end{tabular}
\caption{Experimental results for argument clustering over three-fold cross-validation. 
Each value in the table is the average over three trials.
\#C indicates the final number of clusters.}
\label{tab:clustering}
\end{table*}

We also compared our methods with the two unsupervised methods used in Subtask-B.1 of SemEval-2019 Task 2 \cite{qasemizadeh-2019-semeval}.\footnote{The SemEval-2019 Task 2 dataset is no longer available, as described on its official website; thus, we excluded that dataset from the experiments.}
\newcite{anwar-2019-hhmm} performed group-average clustering by using a negative one-hot encoding feature vector to represent the inbound dependencies of argument words.
\newcite{ribeiro-2019-l2f} applied graph clustering by Chinese whispers \cite{biemann2006} with the average ELMo \cite{peters-2018-deep} embeddings of argument tokens.
We also prepared two baselines: \textbf{Boolean} and \textbf{Dependency-relationship}.
The Boolean method clusters argument instances based on whether they appear before or after the verb. 
For example, in the second example sentence ``[I] will now \underline{donate} [the money] [to charity].'' in Table \ref{tab:examples}, the word ``I'' belongs to the \textsl{before} cluster, while ``the money'' and ``to charity'' belong to the \textsl{after} cluster.
The Dependency-relationship method clusters argument instances based on dependency labels. 
In the case of the same example sentence as above, ``I'' belongs to a cluster indicating a noun subject, ``the money'' belongs to a cluster indicating an object, and ``to charity'' belongs to a cluster indicating an oblique nominal.
We use stanza \cite{qi-etal-2020-stanza} as a dependency parsing tool.\footnote{\url{https://stanfordnlp.github.io/stanza/}}


\paragraph{Metrics}
For evaluation metrics, we used \textsc{Purity} (\textsc{Pu}), \textsc{inverse Purity} (\textsc{iPu}), and their harmonic mean, \textsc{F-score} (\textsc{PiF}) \cite{zhao-2001-criterion}, as well as \textsc{B-cubed Precision} \textsc{(BcP}), \textsc{Recall} (\textsc{BcR}), and their harmonic mean, \textsc{F-score} (\textsc{BcF}) \cite{bagga-1998-entity-based}.

\begin{figure*}[!t]
\centering
\includegraphics[width=\linewidth]{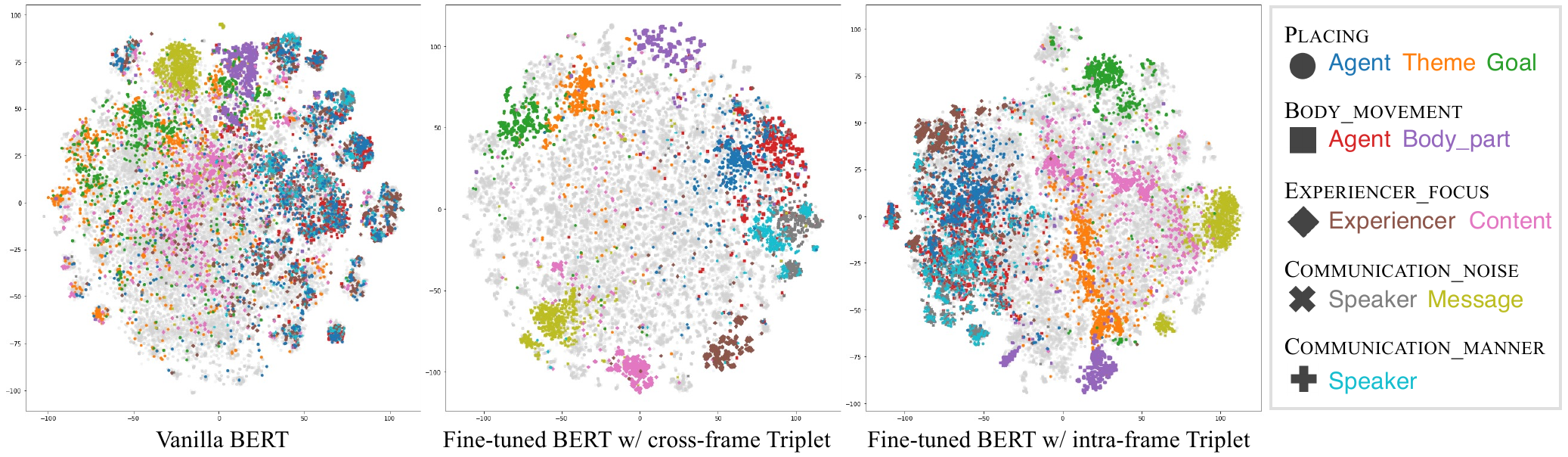}
\caption{2D t-SNE projections of the average embeddings of argument tokens with the Vanilla, cross-frame Triplet, and intra-frame Triplet models. 
The top 10 FEs with the highest numbers of instances are highlighted.}
\label{fig:visualization}
\end{figure*}

\subsection{Results}
Table \ref{tab:clustering} summarizes the experimental results.
The cross-frame and intra-frame methods with the Triplet and ArcFace models showed a remarkable performance improvement compared to those with the Vanilla model.
In particular, the intra-frame method with the Triplet model obtained a high score of 92.5 for \textsc{PiF} and 88.6 for \textsc{BcF}.
Also, while there was no difference between the intra-frame and cross-frame methods with the Vanilla model, we can confirm the efficacy of the intra-frame methods with the fine-tuned models.
There was little difference in scores with the deep metric learning models.
We consider that they achieved similar scores as a result of satisfactory learning because both models learn margin-based distances.

As for the comparison to previous methods, the methods with the Vanilla model underperformed the baseline methods with syntactic features, but our methods with the fine-tuned models outperformed them considerably.
This result also confirms the usefulness of the fine-tuned models through deep metric learning.
Among the previous methods, although the two baselines performed better than the methods in \newcite{anwar-2019-hhmm} and \newcite{ribeiro-2019-l2f}, this was an expected result because the experiment by Anwar et al. showed that the Boolean method obtained higher scores than their method.
Note that our experiment only considered core FEs.
The trends that baselines with syntactic features performed well may not be going to hold in experiments that consider non-core FEs.

We also visualized the embeddings to understand them intuitively.
Figure \ref{fig:visualization} shows a 2D t-SNE projection of the average contextualized embeddings of the argument tokens.
With the Vanilla model, clumps of instances can be seen for each FE, but instances for the same FE are entirely scattered, and the instances for different FEs in the same frame are mixed together.
On the other hand, with the fine-tuned models, the instances are clustered for each FE.
We can see that the instances with the cross-frame Triplet model are tightly grouped by FEs than those with the intra-frame Triplet model.
However, the FEs are still independent of each frame, and it is important to distinguish instances of different FEs in the same frame.
The intra-frame Triplet model distinguishes more instances with different roles in the same frame than the cross-frame Triplet model does, such as instances of \textsf{\small Theme} and \textsf{\small Goal} in the \textsc{Placing} frame.
Furthermore, with the intra-frame Triplet model, we can see instances of similar roles clustered together across frames such as instances of \textsf{\small Speaker} in the \textsc{Communication\_noise} frame and \textsf{\small Agent} in the \textsc{Placing} frame.
These results confirm the usefulness of the fine-tuning of the intra-frame method.

\section{Conclusion}
We have addressed argument clustering for semantic frame induction.
We proposed a method that uses deep metric learning to fine-tune contextualized embedding models and applied the resulting fine-tuned embeddings to perform argument clustering.
We also introduced intra-frame methods that exploit the property that FEs are frame-specific.
Experimental results showed that fine-tuned models with deep metric learning are promising and that intra-frame methods perform quite well.
Especially, the intra-frame method with the Triplet model achieved high scores of 92.5 for \textsc{PiF} and 88.6 for \textsc{BcF}.

Although only core frame elements are covered in this study, it would be ideal to acquire non-core frame element knowledge as well.
Since many non-core frame elements are shared among different frames and are likely to be easier to learn than core frame elements, our methods are expected to achieve competitive performance for non-core frame elements as well.
We would like to confirm it in future work.
The ultimate goal of this research is to automatically build frame knowledge resources from large text corpora.
We will need to merge our method with methods that cluster verbs according to the frames that they evoke \cite{yamada-2021-semantic, yamada-etal-2023-semantic} and predict the positions of argument tokens.
In addition, we will consider how to apply our method to large text corpora.

\section*{Limitations}
As we only used English FrameNet as the dataset for our experiment, it is unclear how well our method would work with other languages or corpora.
However, because the method is neither language- nor corpus-specific, fine-tuning may lead to better results with other datasets. 
Also, the method relies on a semantic frame knowledge resource, and annotation will thus be required if it is applied to languages without such resources.
This study only considers core frame elements and does not show results for non-core frame elements.

\section*{Acknowledgements}
This work was supported by JST FOREST Program, Grant Number JPMJFR216N and JSPS KAKENHI Grant Numbers 21K12012 and 23KJ1052.

\bibliography{acl2023}
\bibliographystyle{acl_natbib}

\appendix
\section{How to Determine Number of Clusters}
 \label{sec:number_of_clusters}
Here, we explain how to determine the number of clusters in cross-frame and intra-frame methods.
In the cross-frame method, it is determined from the ratio of the number of FEs to the number of frames in the development set.

In contrast, the intra-frame method uses criteria across frames because the number of frames is not easy to decide on a frame-by-frame basis.
The termination criterion for clustering is the point at which there are no more cluster pairs for which the distance between clusters is less than a threshold $\theta$ that all frames share. 
The threshold $\theta$ is gradually decreased from a sufficiently large value, and the average number of clusters over all frames is set to a value that is closest to the average number of different FEs in each frame in the development set.

\section{Detailed Settings for Our Methods}
 \label{sec:detail_settings}
Here, we describe the detailed settings, including hyperparameters, of the methods in our experiment.
All embeddings were processed with $l_2$ normalization to match the ArcFace requirement.
In fine-tuning, the batch size was 16, the learning rate was 1e-5, and the number of epochs was 10.
The candidate margins were 0.1, 0.2, 0.5, and 1.0 for the triplet loss and 0.01, 0.02, 0.05, and 0.1 for the ArcFace loss.
The feature scale for ArcFace was 64.
We explored only the margin because \newcite{zhang-2019-adacos} showed that the behaviors of the margin and scale are similar.
The optimization algorithm was AdamW \cite{loshchilov-2017-decoupled}.

In the experiment, the epochs and margins for fine-tuning and the number of clusters for clustering were determined by the development set.
The most plausible model for fine-tuning was determined from ranking similarities to ensure clustering-independent evaluation.
Specifically, we took an argument instance as a query instance; then, we computed the cosine similarity of the embeddings between the query instance and the remaining argument instances, and we evaluated the instances' similarity rankings in descending order.
For a metric, we chose the recall.
It computes the average match rate between true instances, which are instances of the same FE as the query instance, and predicted instances, which are obtained by extracting the same number of top-ranked instances as the number of true instances.
The embedding of the model with the highest score was used for clustering.
\end{document}